\documentclass[11pt]{article}
\pdfoutput=1   
\usepackage[preprint]{acl}

\usepackage{times}
\usepackage{latexsym}
\usepackage[T1]{fontenc}
\usepackage[utf8]{inputenc}
\usepackage{microtype}
\usepackage{inconsolata}
\usepackage{hyperref}
\usepackage{url}
\usepackage{booktabs}
\usepackage{amsfonts}
\usepackage{amsmath}
\usepackage{amssymb}
\usepackage{graphicx}
\graphicspath{{figures/}{../figures/}{./}}
\usepackage{xcolor}
\usepackage{xspace}
\usepackage{fontawesome5}
\usepackage{longtable}
\usepackage{array}
\usepackage{multirow}
\usepackage{tikz}
\usetikzlibrary{arrows.meta, positioning}
\definecolor{detblue}{HTML}{2a78d6}
\definecolor{modorange}{HTML}{eb6834}
\usepackage[most]{tcolorbox}
\definecolor{boxframe}{RGB}{170,170,170}
\definecolor{boxtitlebg}{RGB}{208,208,208}
\tcbset{graybox/.style={
  breakable, enhanced, colback=gray!3, colframe=boxframe,
  colbacktitle=boxtitlebg, coltitle=black, fonttitle=\bfseries\small,
  arc=0.6mm, boxrule=0.4pt, left=2mm, right=2mm, top=1mm, bottom=1mm,
  before skip=3pt, after skip=3pt}}
\newtcolorbox{promptbox}[1][]{graybox, fontupper=\footnotesize\ttfamily, title={#1}}
\newcolumntype{L}[1]{>{\raggedright\arraybackslash}p{#1}}

\newcommand{\sys}{\textsc{factwash}\xspace}

\title{\sys: Catching AI Rewrites That Wash Hearsay into Fact\\
{\large\normalfont Linguistic class predicts where deterministic checking suffices}}

\author{
  Alex Kwon \\
  Independent Researcher \\
  \texttt{ask@collapseindex.org} \\[4pt]
  \href{https://github.com/collapseindex/factwash}{\faGithub~GitHub}
}

\begin{document}
\maketitle

\begin{abstract}
AI systems rewrite information constantly: conversations become stored memories,
documents become answers. The rewrite can keep a claim while washing away what made it
checkable, who said it, how sure they were, when it held. We call that failure
\emph{factwashing}, and release \sys, an open-source write-time gate that catches it
deterministically, with named flags and evidence rather than an LLM judge. Building it
answers a practical question: when does a cheap check suffice, and when do you need a
model? What decides is whether the property has a \emph{bounded surface-cue inventory}.
Explicit negation cues are close to enumerable, so a word list finishes and transfers,
reaching $0.91$ F1 on untuned text. Hedging and attribution have open-ended realizations,
so vocabulary plateaus near half recall, and a one-question LLM \emph{witness} recovers
$+17$ and $+15$ points of \emph{cue-detection} recall at equal precision. Deployed, that
witness may only lower a verdict, so it buys precision rather than coverage. We measure
cue detection on $105{,}596$ independently annotated sentences. A blind-labelled corpus of memory
writes then locates the failure: $55\%$ of bad writes in conversational hearsay, $7\%$ in
business email ($p < 0.001$), so the first deployment question is not which detector to
use but whether the failure occurs at all. On unmodified \texttt{mem0} $2.0.7$, the gate
flags $5$ of $8$ hedged-hearsay writes.
\end{abstract}

\begin{figure*}[t]
\centering
\resizebox{\textwidth}{!}{%
\begin{tikzpicture}[
  font=\small,
  box/.style={draw=gray!70, fill=gray!6, rounded corners=2pt, align=center,
              inner sep=5pt, minimum height=8mm},
  det/.style={box, draw=detblue!80},
  opt/.style={box, draw=modorange, dashed},
  lbl/.style={font=\footnotesize, text=black!60},
  arr/.style={-{Stealth[length=2.2mm]}, gray!80, thick},
  optarr/.style={-{Stealth[length=2.2mm]}, modorange, thick, dashed},
  node distance=6mm and 9mm]

\node[box] (src) {source\\context};
\node[box, below=3mm of src] (mem) {candidate\\memory};
\node[det, right=of src, yshift=-7mm] (align) {align claims\\ \footnotesize content-word overlap};
\node[det, right=of align] (checks) {\textbf{seven checks}\\
  \footnotesize polarity \(\cdot\) condition \(\cdot\) temporal\\
  \footnotesize hedge \(\cdot\) attribution\\
  \footnotesize brittle \(\cdot\) truncated};
\node[det, right=of checks] (flags) {flags\\ \footnotesize evidence + citation};
\node[det, right=of flags] (policy) {verdict policy\\ \footnotesize hard-coded};
\node[box, right=of policy, align=left] (verd)
  {\textsc{pass}\\ \textsc{pass\_w\_flags}\\ \textsc{rewrite}\\ \textsc{reject}};
\node[box, below=10mm of verd, align=center, inner xsep=8pt] (unch)
  {\textsc{uncheckable}\\ \footnotesize never a pass};

\node[opt, below=7mm of checks] (wit) {\textbf{witness} (optional)\\
  \footnotesize one sentence, one question\\
  \footnotesize markers validated against text\\
  \footnotesize can only \emph{lower} a verdict};
\node[opt, below=7mm of policy] (fix) {\textbf{fixer} (optional)\\
  \footnotesize LLM rewrite, re-gated};

\draw[arr] (src.east) -- (align.150);
\draw[arr] (mem.east) -- (align.210);
\draw[arr] (align) -- (checks);
\draw[arr] (checks) -- (flags);
\draw[arr] (flags) -- (policy);
\draw[arr] (policy) -- (verd);
\draw[arr, shorten >=4pt] (align.south) |- (unch.180)
  node[lbl, pos=0.28, right] {no match};
\draw[optarr] (flags.south) |- (wit.east);
\draw[optarr] (wit.north) -- (checks.south)
  node[lbl, pos=0.5, right] {veto};
\draw[optarr] (verd.190) -| (fix.north)
  node[lbl, pos=0.85, right] {\textsc{rewrite}};
\draw[optarr] (fix.west) -| (align.south east)
  node[lbl, pos=0.15, above] {re-gate};
\end{tikzpicture}}
\caption{\textbf{The \sys gate in one view.} Solid blue path: deterministic, offline,
zero dependencies; every flag carries its evidence and a citation. A write no check can
align with its source is \textsc{uncheckable}, never a pass. Dashed orange: the two
optional model components, both subordinate to the deterministic path: the witness
answers one question about one sentence and can only lower a verdict; the fixer's
rewrite must re-pass the same gate. \texttt{factwash.inspect()} re-reports the same
flags as typed changes (\textsc{dropped} attribution, \textsc{strengthened} certainty,
\dots) with undetected types declared.}
\label{fig:pipeline}
\end{figure*}
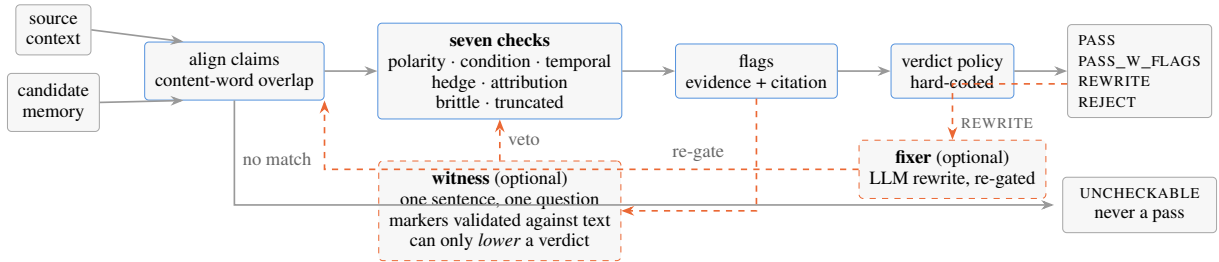

\section{Introduction}

An agent is told: ``someone said Alice was promoted this morning.'' The memory system
stores: ``Alice was elevated to administrator status on August 3, 2026.'' A later step
reads that and grants Alice access she never had. Nothing was hallucinated. The fact
survived; what was lost was that it was hearsay.

We call this failure \emph{factwashing}: a rewrite that preserves a claim while washing
away its epistemic standing. A memory can become \textbf{uncorrectable}, when the basis
for a claim is dropped and nobody can later check it, or \textbf{falsely confident},
when the hedge or the attribution is stripped and a tentative claim reads as settled.
Neither is hallucination: a factuality metric scores both as correct.

The check compares the stored text against its source; the question is what the
comparison is made of: a free, auditable word list, or an LLM call that must be
trusted. We find the
choice is not a matter of taste. It is predicted by the linguistic class of the property
being checked. Negation is a \emph{closed class}: English has a small, stable set of
ways to say ``not,'' so a list can be finished, and a finished list transfers to text it
was never built for. Hedging and attribution are \emph{open classes}: there is no
complete list of ways to signal ``I am not sure'' or ``someone told me,'' so no list can
be finished, and every list stalls at what its author thought of.

The distinction is testable, and it held on all four properties we could test against
annotation nobody here wrote: the witness gain lands on the two open classes and no
other (\S\ref{sec:external}).

\paragraph{This preprint documents four things.}
(1)~\sys: a zero-dependency write-time gate (\texttt{pip install factwash}) with seven
deterministic checks, evidence-validated model components, drift tracing, and a
benchmark scoring a system's laundering rate (\S\ref{sec:tool}).
(2)~The \textbf{design rule}: closed-class properties are checkable by transferable
word lists, open-class ones are not, and only there does a model pay
(\S\ref{sec:external}).
(3)~\textbf{External validation} on $105{,}596 + 4{,}800$ independently annotated
sentences, with a lexicon-vs-witness head-to-head (\S\ref{sec:external}).
(4)~A \textbf{boundary}: factwashing dominates hearsay's bad writes and is rare in
business email (\S\ref{sec:writes}).

\section{Related Work}
\label{sec:related}

\paragraph{Why memory systems compress, and store.}
Compression is forced, not chosen: attention costs grow with context, so
\citet{jiang2023llmlingua} drop low-information tokens from prompts, and for dialogue
the same pressure produces summarization, as in the running summary of
\citet{wang2025recursive}. Note what those methods optimise: task accuracy, latency and
token count. A compression that preserves the answer while dropping who said it and how
sure they were scores well on all three. Agent memory then makes the loss durable by
\emph{storing} the result. \citet{packer2023memgpt} manage a working memory against an
archive with the model deciding what to write; \citet{chhikara2025mem0} extract salient
facts and consolidate them. Both treat the write as summarization and tune it for cost
and retrieval quality; neither checks whether the write preserved what made the claim
checkable, and a stored memory outlives the conversation that could have corrected it.

\paragraph{What compression costs.}
Two recent results motivate the checks. \citet{kwon2026reclaim} shows a memory that
keeps a conclusion and drops the values it came from can leave a system worse off than
having no memory, because the wrong answer survives and the means to fix it does not.
\citet{kwon2026mc} shows the stance case: a hedged remark stored as a flat assertion is
obeyed like a verified fact, the agent keying on the confidence of the phrasing rather
than on the source. We take those as the failure modes to detect, and ask when detecting
them needs a model.

\paragraph{Cue annotation, and what it is not for.}
Detecting hedges is a solved annotation problem: \citet{vincze2008bioscope} annotate
speculation and negation cues with scopes, \citet{farkas2010conll} made cue and scope
detection a shared task, and \citet{szarvas2012cross} extend it across genres, studying
exactly the transfer question we care about. Attribution has the same shape in
\citet{pareti2016parc} and \citet{newell2018polnear}, the latter token-level over
political news; we score against the latter, since the former sits on licensed newswire.
We reuse this annotation rather than build our own, on a task none of it was built for.

\paragraph{Summarization faithfulness.}
The closest analogue asks whether a compressed text still says what its source said:
\citet{pagnoni2021frank} collect typed human error labels on generated summaries and
\citet{tang2023aggrefact} aggregate nine such datasets. We use FRANK as an external
check on the whole gate, but it cannot substitute for the task. Its typology is
dominated by hallucination, where the summary states something the source never
contained; our failure is the opposite, the claim right and its standing gone, which a
factuality metric scores as correct.

\section{What a memory write loses}
\label{sec:problem}

The unit we work on is a pair: the conversation a memory was written from, and the memory
itself. We do not ask whether the memory is true, but what it did to the source's claim,
which is answerable by comparing the two texts.

Table~\ref{tab:checks} lists the seven checks, five about stance and scope and two about
arithmetic. The class column is a prediction made before any external evaluation, and it
is what \S\ref{sec:external} tests: a closed-class check should work on text it was never
built for, an open-class one should not.

\begin{table}[t]
\centering\small
\begin{tabular}{@{}lll@{}}
\toprule
Check & What it catches & Class \\
\midrule
\textsc{polarity} & source denied it, memory asserts & closed \\
\textsc{condition} & ``if X'' dropped & closed \\
\textsc{hedge} & source hedged, memory flat & open \\
\textsc{attribution} & who said it is gone & open \\
\textsc{temporal} & source dated it, memory undated & closed \\
\textsc{brittle} & conclusion kept, inputs gone & n/a \\
\textsc{truncated} & total no longer recomputes & n/a \\
\bottomrule
\end{tabular}
\caption{\textbf{The seven checks, and the class of each.} The class column is the
paper's prediction, assigned before the external evaluation: closed-class properties
should transfer across domains, open-class ones should not.}
\label{tab:checks}
\end{table}

Each check needs to know which part of the source a stored sentence came from: we match
on content-word overlap and take the best-scoring source sentence. How much context that
sentence carries differs per check and was measured rather than assumed;
Appendix~\ref{app:corpora} gives the windows and thresholds.

If nothing in the source matches a stored sentence, the checks report \textsc{uncheckable}
rather than passing it. An unchecked write reported as verified would be the same mistake
the tool exists to catch.

\section{The released tool}
\label{sec:tool}

\sys is a Python package (\texttt{pip install factwash}, Apache-2.0, zero runtime
dependencies) exposing the seven checks as a gate (Figure~\ref{fig:pipeline}):

\begin{promptbox}
report = factwash.check(source, memory)\\
report.verdict\ \ \# PASS | PASS\_WITH\_FLAGS | REWRITE | REJECT | UNCHECKABLE
\end{promptbox}

The verdict policy is hard-coded, not scored: uncorrectable failures reject, fixable
stance failures rewrite, and a write no check could align with the source is
\textsc{uncheckable}, never passed. A \texttt{wrap()} adapter gates an existing
\texttt{mem0} store, and the optional witness of \S\ref{sec:witness} attaches as a
callable that can only lower a verdict.

\paragraph{Typed changes.} A second surface, \texttt{factwash.inspect()}, re-reports the
same checks as typed source-to-output changes: \textsc{dropped} attribution,
\textsc{strengthened} certainty, \textsc{reversed} polarity, \textsc{dropped} temporal
scope. Change types with no detector (\textsc{broadened}, \textsc{weakened}) are
declared on every report as \texttt{not\_checked} rather than silently absent, because a
report that lists only what it found reads as ``nothing else happened.'' An optional
units detector \citep{quantulum3} extends this with value-keyed unit drift: ``1.2
million dollars'' stored as ``1.2 million euros'' is caught as a \textsc{changed} unit
even though every stance check passes, a failure the gate structurally cannot see
because no hedge, attribution, or negation moved.

\paragraph{The other direction.} The seven checks ask whether what was in the source
survived. An optional \textsc{added} detector asks the reverse, whether what is in the
memory was ever there, which is the failure that dominated our labelled corpus
($27$ of $29$ bad writes, \S\ref{sec:baserate}) and that the checks structurally cannot
see. It follows the witness architecture with one addition: the model returns
supported / unsupported / cannot-tell for one memory sentence against the source, and
\emph{both} answers must quote the source verbatim, since an ``unsupported'' verdict
must cite the closest source text to prove the model read before claiming absence.
Quotes are validated; a reply that cannot point produces nothing, and sentences the
backend could not establish are declared rather than passed. Measured on the same
blind corpus, against the $27$ fabricated-or-inferred positives and $65$ clean
negatives: precision $0.57$ ($8$ of $14$ flagged), recall $0.30$ ($8$ of $27$), coverage
$96.9\%$. Denominators that small carry wide intervals ($95\%$: $[0.33, 0.79]$ and
$[0.16, 0.48]$). The gain is additive, since the gate catches none of this class by
construction, and the prompt was not tuned against the corpus it scores on.

\paragraph{Chains, not just writes.} Memories are rewritten repeatedly, so
\texttt{factwash.drift()} traces a version history, reporting per-hop changes and
attributing each end-to-end loss to the hop where it happened. On three chains probed
before the feature was built, two behaviours appeared. Within lexicon coverage the gate
\emph{composed}: the hop dropping a class's last cue fired, so those chains could not
launder gradually past per-write gating. That generalises as far as cue presence does,
which is not a proof. The paraphrase chain escaped instead by \emph{losing
checkability}, its hops going \textsc{uncheckable}, which the report surfaces as the
finding it is. Both point the same way: multi-hop danger concentrates where single-hop
danger already lived, outside the lexicon and past alignment.

\paragraph{Scoring a memory system.} \texttt{factwash bench} inverts the gate into a
scorer: given the writes a memory system produced, it reports the share of checkable
writes the gate flags, \emph{beside} the share it could not align and the share of
sources the system stored nothing for. Reporting the three together is deliberate,
because the flag rate alone is gameable: a system whose writes cannot be aligned to
their sources, or that writes rarely, offers fewer chances to be flagged, so a low score
can be evasion rather than cleanliness. The flag rate is also not a verified laundering
count and errs both ways, since bounded recall hides cases while imperfect precision
(about one flag in four is a false alarm on real output) means it is not a floor.
\textsc{uncheckable} writes never enter the denominator, and every report carries its
stimulus-set identifier, since \S\ref{sec:baserate} shows base rates are
domain-dependent.

\paragraph{The deployment contract, in one paragraph.} Gate stores that ingest human
conversation and feed decisions; Table~\ref{tab:baserate} is the decision chart, and on
clean-factual pipelines the gate is mostly idle. Operationally: \textsc{reject} means do
not store, \textsc{rewrite} means store the fixed text or hold for review,
\textsc{uncheckable} means keep but log as unverified. About one flag in four is a false
alarm on real output ($0.73$ precision, domain-dependent), so the default posture is
review rather than block, and the witness is worth enabling when halving that is worth a
cent per hundred writes. Nothing leaves the machine unless the witness or fixer is
enabled, and then one sentence per call.

\paragraph{Claims stay tethered to behaviour.} Every figure published in the project
README is recomputed from the shipped corpora by a test that fails if the text drifts
from the measurement, and that guard is itself negative-tested. The same discipline
produced Appendix~\ref{app:claims}: the project has published wrong numbers twice by
drift, and treating documentation as an asserted artifact is the countermeasure.

\section{External evaluation}
\label{sec:external}

A tool evaluated only on a corpus its author wrote is a self-portrait. The author picks the
examples, writes the labels, and then tunes against both. We built such a corpus first, and
it flattered the tool four separate times before we stopped trusting it
(\S\ref{sec:limits}).

So the checks are also scored against corpora annotated by other people, for other
purposes, before this work existed. None of them was built for memory integrity, and none
of their annotators had any stake in these numbers.

\subsection{Corpora}
We use three annotated corpora for cue detection and one for the whole gate.
\citet{vincze2008bioscope} and \citet{szarvas2012cross} supply speculation and negation
cues over biomedical abstracts, full papers, encyclopedic text and news; together they give
$57{,}891$ sentences. \citet{newell2018polnear} supply attribution as source, cue and
content spans over $1{,}008$ political news articles, or $47{,}705$ sentences. That is
$105{,}596$ sentences in total. \citet{pagnoni2021frank} supply typed human error labels on
generated summaries, which we use in \S\ref{sec:frank}.

Two of the five stance checks have no external gold here: none of these corpora annotate
temporal scope, so the closed-class transfer prediction for \textsc{temporal} is untested
in this section (its false-positive rate is bounded on the should-pass corpus instead),
and the two arithmetic checks have no analogue in cue annotation at all.

The Szeged annotation is the more useful of the two cue sets because its subtypes fall
on our distinctions rather than across them: \emph{modal} and \emph{doxastic} are
hedging, \emph{condition} is our conditional check, and \emph{investigation} (``we
examined whether X'') is research framing we exclude and count.

\paragraph{Discipline.} Documents are split in half; terms were mined from \emph{dev}
under a rule fixed before we looked, and every number below is from the disjoint
\emph{test} half. PolNeAR's own split is used as shipped, which is better than ours
because someone with no stake in the result drew the line.
Appendix~\ref{app:corpora} gives the rule, the thresholds and one exclusion that looked
like a result until it was traced to a redacted corpus release.

\subsection{Closed-class checks transfer; open-class ones do not}
\label{sec:split}

\begin{table}[t]
\centering\small
\begin{tabular}{@{}lccccc@{}}
\toprule
Detector & P & R & F1 & Class \\
\midrule
Negation & $0.89$ & $0.94$ & $\mathbf{0.91}$ & closed \\
Conditionals & $0.61$ & $0.67$ & $0.64$ & closed \\
Hedges & $0.89$ & $0.66$ & $0.76$ & open \\
Attribution & $0.91$ & $0.49$ & $0.63$ & open \\
\bottomrule
\end{tabular}
\caption{\textbf{Held-out cue detection against expert annotation.} Documents are split
so that no sentence from a tuned-on document is scored. The two open classes sit at high
precision and roughly half recall: the word list finds what it knows and cannot be made
to know the rest.}
\label{tab:external}
\end{table}

Table~\ref{tab:external} splits the way the prediction says it should. Negation reaches
$0.91$ F1 on domains it was never tuned on. The two open classes reach high precision and
about half recall.

\begin{figure}[t]
\centering
\includegraphics[width=\columnwidth]{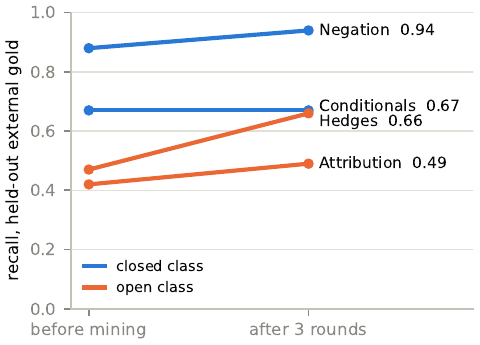}
\caption{\textbf{Mining closes a closed class and stalls on the open ones.} Held-out
recall before and after three rounds of lexicon mining from external corpora, under one
rule fixed in advance. Negation reaches $0.94$ and stops because nothing further clears
the rule; conditionals gain nothing from vocabulary at all (their improvement was
precision, from narrowing); hedging and attribution rise and stall, and the misses that
remain are not a shorter list of the same kind.}
\label{fig:mining}
\end{figure}

The point is not that negation is easier. It is that negation is \emph{finishable}
(Figure~\ref{fig:mining}). When
we mined the dev half for terms the negation list was missing, three cleared the bar and
recall went from $0.88$ to $0.94$; after that there was nothing left that met the rule.
The list is close to complete because the thing it describes is close to complete. The
same mining on hedging moved recall from $0.47$ to $0.66$, and on attribution from $0.42$
to $0.49$, and in both cases the misses that remain are not a shorter list of the same
kind. They are an open set.

Conditionals are the instructive case. No vocabulary candidate cleared the bar at all, and
the gain there came from \emph{removing} rather than adding: ``subject to'' usually means
susceptibility (``subject to change''), ``assuming'' is often a plain verb (``assuming
command''), and an ``if'' that means ``whether'' (``we tested if X held'') introduces a
complement, not a condition. Precision improved from $0.55$ to $0.61$ by narrowing three
terms. A closed class can be over-covered as well as finished; an open class can be
neither.

\paragraph{What is and is not being claimed.} That closed classes have fewer members
than open ones is a fact about English, not a finding. The claim is the engineering
consequence, which does not follow from the definition and is not usually tested: class
membership tells you \emph{in advance} whether adding vocabulary will repay the effort,
and therefore where a model is worth paying for. Two results give that prediction teeth.
The mining rule was fixed before we looked and applied identically to every class, and
it closed negation while failing to close hedging or attribution across three rounds;
and the witness gain appears on the open classes and is unavailable on the closed one,
because nothing is left there to win. A survey of list sizes would show neither.

\subsection{A word list cannot be finished}
The clearest evidence that this is a property of the class rather than a lack of effort is
what happens when you try harder.

\begin{figure}[t]
\centering
\includegraphics[width=\columnwidth]{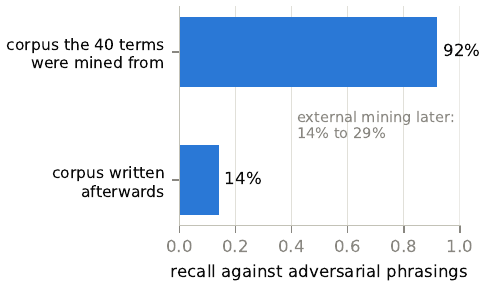}
\caption{\textbf{The lexicon memorises the corpus it can see.} Recall of the same forty
added terms against the adversarial corpus they were mined from versus a corpus written
afterwards in the same spirit. Three later rounds of mining from external corpora moved
the held-out number from $14\%$ to $29\%$: two cases out of fourteen.}
\label{fig:memorise}
\end{figure}

We wrote an adversarial set of hedged and attributed phrasings deliberately outside the
list, and added forty terms to catch them. On the corpus we could see, recall went from
$25\%$ to $92\%$. On a second set written afterwards, in the same spirit but not looked at
during the additions, it was $14\%$ (Figure~\ref{fig:memorise}). The list had memorised
the visible corpus.

Three subsequent rounds of mining from external corpora moved that held-out number from
$14\%$ to $29\%$: two more cases out of fourteen. The one that landed is instructive.
``Sources say'' is now caught, because the mining finally added the stem \emph{say} to a
list that already had \emph{said} and \emph{says}. That is a real gap and fixing it was
worth doing. It is also exactly what an open class looks like from the inside: every round
of work buys a couple of specific phrasings and never the category.

\section{Where the model earns its cost}
\label{sec:witness}

If the gap is really about open classes, swapping the word list for something that does
not depend on one should close it, and only there. That is this section's test.

\paragraph{The witness perceives; it does not judge.}
We give a model one sentence and ask whether it hedges and whether it attributes. It
returns two booleans and the exact words that made it answer yes; it never sees the other
sentence, never compares, and never returns a verdict, so the comparison rule and verdict
policy stay in code and the detector is the only thing that changes. Every quoted marker
is checked against the sentence it came from, and an answer citing words that are not
there is discarded, leaving the deterministic verdict standing. That matters more than it
looks: a witness that cannot point at the text is guessing, and a guess that reaches a
verdict is an LLM judge with extra steps.

\begin{table}[t]
\centering\small
\begin{tabular}{@{}llccc@{}}
\toprule
Task & Detector & P & R & F1 \\
\midrule
\multirow{2}{*}{Hedges} & word list & $0.97$ & $0.62$ & $0.76$ \\
 & witness & $0.96$ & $\mathbf{0.79}$ & $\mathbf{0.87}$ \\
\midrule
\multirow{2}{*}{Attribution} & word list & $0.85$ & $0.45$ & $0.59$ \\
 & witness & $0.87$ & $\mathbf{0.60}$ & $\mathbf{0.71}$ \\
\bottomrule
\end{tabular}
\caption{\textbf{Witness versus word list on the same expert gold}, 200 held-out
sentences per task, stratified. The witness gains recall at equal precision, on exactly
the two open classes. Unusable replies are counted as misses.}
\label{tab:witness}
\end{table}

We score both detectors on the same $200$ held-out sentences per task, sampled half
positive and half negative so that answering ``no'' to everything cannot look good. The
word-list rows therefore differ from Table~\ref{tab:external}, which uses the full test
half at its natural class balance (the subsample shifts hedge precision $0.89 \to 0.97$,
recall $0.66 \to 0.62$); both detectors face the identical subsample, so the comparison
is unaffected. The witness runs through the shipped code path, span validation included.
When it returns something unusable we count it as a miss rather than skipping it, because
that is what the gate does with it, so the numbers in Table~\ref{tab:witness} understate
what the model
perceived.

The witness gains $17$ points of recall on hedging and $15$ on attribution, at
precision unchanged within noise. That is a large gain and it is not the interesting part.
The interesting part is where it appears. It appears on the two open classes, which is
where the prediction says a list cannot be finished, and there is nothing for it to win on
negation, where the list already found what there was to find.

One reading must be blocked, because it is the natural one: this is
\emph{detector-level} recall on isolated sentences, not gate recall on writes. The
shipped witness may only lower a verdict, so enabling it buys precision, not coverage
(\S\ref{sec:limits} reports what happened when we let it raise verdicts). The result
says the open-class ceiling belongs to word lists rather than to the task, which is why
a model belongs in the architecture; it does not say installing one buys $17$ points.

\paragraph{Cost.} All $400$ calls cost \$$0.157$ on a small model
\citep[claude-haiku-4.5;][]{anthropic2025haiku}. In the shipped
configuration the witness runs only on writes the deterministic layer already flagged,
which is about one cent per hundred writes.

\subsection{Against the obvious baseline: a direct LLM judge}
\label{sec:judge}
The question every reader asks is why not simply hand the pair to a model. We did, with
the same rubric the human labeller used (``would this memory mislead someone reading it
later?''), scored against the same labels, on the two sets that can support the
comparison (Table~\ref{tab:judge}).

\begin{table}[t]
\centering\small
\begin{tabular}{@{}llcc@{}}
\toprule
Set & Detector & P & R \\
\midrule
\multirow{3}{*}{\shortstack[l]{Real writes\\ ($44$, $20$ pos.)}}
 & gate & $0.73$ & $\mathbf{0.95}$ \\
 & judge, haiku-4.5 & $1.00$ & $0.30$ \\
 & judge, sonnet-5 & $0.91$ & $0.50$ \\
\midrule
\multirow{2}{*}{\shortstack[l]{Adversarial\\ ($46$, $26$ pos.)}}
 & gate & $1.00$ & $0.58$ \\
 & judge, haiku-4.5 & $1.00$ & $\mathbf{0.92}$ \\
\bottomrule
\end{tabular}
\caption{\textbf{A pairwise LLM judge versus the gate}, same gold, same rubric. The
judge wins where phrasing is unfamiliar and loses where writes are real. Models are
claude-haiku-4.5 \citep{anthropic2025haiku} and claude-sonnet-5
\citep{anthropic2025sonnet}.}
\label{tab:judge}
\end{table}

The split is sharp. On adversarial phrasing the judge nearly doubles the gate's recall,
which is the open-class result of \S\ref{sec:external} arriving by another route. On
real extractor output it inverts: near-perfect precision, and it misses fourteen of the
twenty writes the labeller flagged. Scaling the model from haiku-4.5 to sonnet-5
narrows the gap without closing it.

The reason is in the judge's own explanations, and it is the paper's thesis restated by
the system meant to detect it. Its misses say ``the memory accurately captures the
\emph{core fact}'' and ``preserves the key information'': asked whether a memory would
mislead, the judge checks whether the \emph{claim} survived, finds that it did, and
passes. It catches laundering when the source is flagrantly marked (``rumor has it'',
``might be'') and passes it when the write reads plausible, which is the wrong direction
for a gate, because plausible writes are the ones that get acted on.

Two honest qualifications. This is one rubric and two models, and a differently-worded
prompt may do better; the comparison bounds the naive baseline, not every possible
judge. And the judge volunteered a source-grounded quote on $89$ of $90$ items, so the
case for validating evidence is that verdicts must be \emph{required} to point at text,
not that models are unable to.

\subsection{The whole gate against human error labels}
\label{sec:frank}
Scoring the gate rather than its detectors needs pairs, and \citet{pagnoni2021frank} has
them. The headline is unflattering and structural: on $4{,}800$ generated sentences the
gate blocks $37.5\%$ of those all three annotators called clean, because FRANK's errors
are mostly hallucination, which these checks cannot see, and because a news summary that
drops ``according to the AP'' is doing its job where a memory that drops it is not. It
was still worth running: it found two defects no local corpus could, and fixing them cut
the clean-sentence rate from $65\%$ to $37.5\%$ while \emph{raising} agreement on the one
error type we target. Appendix~\ref{app:frank} gives the defects and the numbers.

\section{Memory writes}
\label{sec:writes}

Everything so far scores detectors against annotation. This section asks the question the
tool exists for: on actual memory writes, does the failure occur, and does the gate catch
it? The first question turns out to govern the second: where the failure is rare, catching
it cannot even be measured, and this section reports that boundary rather than a number
without a measurement behind it.

\subsection{A labelled corpus of real writes}
No public corpus of real memory writes exists at usable scale. \citet{packer2023memgpt}
publish agent traces, but they are overwhelmingly retrieval: after de-duplication they
contain $13$ writes, and zero of one documented write operation. So we built one. Source
conversations are Enron email threads \citep{klimt2004enron} that carry quoted or
forwarded content, selected on that structural property alone. Selecting on hedging
vocabulary would have built a pool out of what the word lists already see. Memories are
produced by a generic extraction prompt of the kind a memory system actually uses,
yielding $1{,}415$ candidate writes.

\paragraph{Protocol.} Labelling is blind and stratified, and the sampling frame is worth
stating exactly because the rates invite misreading. The gate split the $1{,}415$
candidates into $406$ flagged and $1{,}009$ passed writes. A stratified session of $300$
was drawn from that pool: $150$ per stratum, which is $37\%$ of the flagged writes and
$15\%$ of the passed ones. Labelling then covered $101$ of those $300$ before analysis.
The labeller sees the source and the memory and nothing else: no verdict, no flags, no
indication of which stratum an item came from, and the two strata are interleaved so
position carries no signal. Every count is scaled by the inverse of its stratum's
pool-level rate, so precision is stable while recall is an estimate with a much wider
interval. \textsc{ambiguous} is a first-class label, excluded from both
figures and reported separately rather than resolved toward whichever answer helps. Of
$101$ labelled writes, $94$ were usable, $4$ ambiguous and $3$ malformed.
Self-agreement, from a blind second pass over $40$ items re-served in fresh order:
$70\%$ raw ($28/40$), Cohen's $\kappa = 0.47$ \citep{cohen1960kappa} over the four
labels; restricted to
flag/pass decisions, $77\%$ ($27/35$), $\kappa = 0.55$. Four of the twelve
disagreements involve the \textsc{ambiguous} boundary. The second pass was also
stricter, flagging six items the first pass had passed against two flips the other
way, so the two passes disagree about magnitude in a consistent direction rather than
symmetrically.

\subsection{Where the failure lives}
\label{sec:baserate}
The first thing the corpus said was not about the gate. Of $29$ writes labelled bad, $27$
were wrong, invented or inferred claims, and only $2$ were the loss of stance or scope
that these checks target. A memory reading ``Lynn works with Steve in logistics'' came
from an email \emph{asking} Lynn and Steve whether logistics could build a report; the
relationship is fabricated. Nothing was hedged away. The claim is simply false.

That is a fact about business email, not about memory writes in general, and the
difference is large. Running the same mechanism question over $20$ bad writes from
conversational hearsay --- the setting \citet{kwon2026mc} constructed --- gives a very
different profile (Table~\ref{tab:baserate}).

\begin{table}[t]
\centering\small
\begin{tabular}{@{}lccc@{}}
\toprule
Source domain & bad writes & stance loss & rate \\
\midrule
Conversational hearsay & $20$ & $11$ & $\mathbf{55\%}$ \\
Business email & $29$ & $2$ & $7\%$ \\
\bottomrule
\end{tabular}
\caption{\textbf{The targeted failure is domain-dependent.} Bad writes classified by
mechanism; ``stance loss'' means a hedge, attribution, negation or scope was dropped
rather than the claim itself being wrong. Fisher exact, $p = 0.00053$.}
\label{tab:baserate}
\end{table}

Three things bound this. The conversational sources were \emph{constructed} to contain
hearsay, so $55\%$ is an upper bound for that setting and not an estimate of natural
conversation. The two corpora differ in more than domain: the email set was sampled and
labelled blind, the conversational set exhaustively and earlier, so the direction is solid
and the magnitude is not a clean effect size. And even in a corpus built to contain
laundering, $9$ of $20$ bad writes were out of scope --- extractors fail in ways beyond
stance loss wherever you look.

\subsection{What that means for the gate}
On the email corpus the gate reaches $0.34$ precision: of the writes it flagged, about one
in three was a write the labeller also called bad. Read alongside
Table~\ref{tab:baserate}, that number is mostly a base-rate result rather than a detector
result. The gate fires on dropped stance tokens, and in this domain dropped stance tokens
are usually harmless, because the claims they attach to were not contested in the first
place.

We do not report a recall figure on the targeted failures for this corpus. Restricted to
in-scope failures the denominator is $2$, and any ratio computed from it would be a number
without a measurement behind it.

\subsection{End to end}
Where the failure does live, the consequence is concrete. Two independent \texttt{mem0}
stores receive the same hearsay and one is wrapped by the gate; an access-control agent
is then asked to grant a resource the subject is not entitled to. The naked store
consolidates ``someone said she was elevated'' into a dated assertion and the agent
grants; the gated store preserves the attribution and the agent escalates. Nothing is
stubbed, including extraction and embeddings.

A demonstration is not a measurement, and extraction is sampled: the same stimulus made
the store keep \emph{nothing at all} in one run and produced our sharpest laundering
example in the scored run below. Variance of that size is itself the argument for
scoring a system over a stimulus set rather than arguing from one example.

\subsection{Scoring a production memory system}
\label{sec:mem0}
The bench turns the gate on unmodified production software. We ran \texttt{mem0}
$2.0.7$ with its own extraction model over a fixed $15$-source stimulus set: ten
hedged-hearsay sources and five confidently-sourced controls, scored as two separate
runs because averaging them would bury the base-rate result of \S\ref{sec:baserate}.

On the hearsay sources \textbf{the gate flags $5$ of $8$ writes} ($62\%$, $95\%$
interval $[0.31, 0.86]$); two more produced no write at all, and abstention is reported
rather than counted as a pass. On the confident controls, one
write of five is flagged, and it is this paper's own documented false positive appearing
in the wild: ``per the IAM system of record'' trips the ported \emph{record} cue
(\S\ref{sec:limits}). That control is how to read the hearsay number: a flag is not a
conviction. The stored text carries the result
better than the rate does. ``Rumor has it Alice now has admin access after the reorg''
was stored as ``Alice was promoted to admin around late July or early August 2026 and
now has admin access'': the hearsay is gone, and a date that was never in the source
has appeared.

Running the \textsc{added} detector over the same writes flags $8$ of $13$, and the
composition of that number is the more useful finding: five are timestamp resolution
(``yesterday'' becoming an absolute date), two are genuine invention where the source
carried no time reference at all, and one is an attribution shift (``reportedly''
becoming ``user reports''). So the dominant false-positive class for \textsc{added} on
a real memory system is date resolution, which is a calibration fact anyone gating on
it needs before they turn it on. These are single-run figures on one stimulus set and
one extraction model, and the caveat that scores are comparable only within a stimulus
set applies to them first.

\section{What it does not catch}
\label{sec:limits}

\paragraph{The deterministic gate has bounded recall, by construction.}
Against phrasing outside its lists it catches $29\%$. This is the paper's own claim turned
on its own tool: the properties it checks are open classes, so no list finishes, and ours
has not either. It is a reason to use the gate where a false alarm costs more than a miss,
and not where you need coverage.

\paragraph{The ceiling is the lexicon, not the matcher, and we checked.} Substring
matching is the obvious suspect for that bounded recall, and embedding alignment the
obvious fix. We attributed every known miss before building anything: of twelve, ten are
detector losses on correctly aligned sentences (``overheard'', ``scuttlebutt'', ``my
sense is''), one is the adversarial item written to defeat substring matching, and one
an inferred claim no matcher can reach. A better matcher recovers at most one of those
twelve, so the claim that the remaining misses need semantics rather than vocabulary
survives an attack on its own infrastructure, at $n=12$.

\paragraph{Turning the witness up does not help, and we measured that.}
The shipped witness can only lower a verdict, so it buys precision and cannot raise recall.
The obvious next move is to let it raise verdicts too, on writes the gate passed. On the
adversarial corpus that reaches $93\%$ recall, which we called a pending improvement until
we measured it. On real labelled writes it gains nothing: recall unchanged,
precision down $7$ points, and all three verdicts it raised were wrong.

The reason generalises: the deterministic layer already catches most of what is catchable
on real output, so what is left for a model to adjudicate is disproportionately what the
model gets wrong. A cascade that escalates where the errors are not spends money to lose
precision.

\paragraph{One failure mode needs ontology, not vocabulary.}
``Alice can access the test server'' stored as ``Alice has server access'' broadens a
permission, and the obvious signal, a dropped modifier on a retained noun, fires on $86$
of $89$ writes that should pass, because ordinary compression drops modifiers constantly.
Separating broadening from summarising means knowing a test server is a kind of server:
world knowledge, not word knowledge, so neither a list nor a witness as posed here.

\section{Conclusion}

Whether you need a model in the loop is not a matter of taste. For a closed-class
property a word list can be finished and transfers to text it was never built for; for
an open-class property no list finishes, and that gap is what a model closes. We found
this building a memory gate, but the argument is not about memory: it applies wherever a
cheap check is weighed against an expensive one, and says which you need before you pay.

\section*{Limitations}

This section is about the measurements rather than the tool: what the gate cannot catch
is \S\ref{sec:limits}, and what should make a reader discount the figures is here.

\paragraph{The rule is about cue inventories, and rests on four properties in one
language.} Two bounds belong on it. First, scope: it predicted transferability on
negation, conditionals, hedging and attribution, in English, and one of those
(conditionals, $0.64$ F1) is handled only moderately and held up by a narrowing argument
rather than a strong number, while \textsc{temporal} has no external gold here at all.
Second, and more important, what the corpora annotate is \emph{cues}. Negation as a
semantic phenomenon is not closed: it surfaces through \emph{lack}, \emph{fail to},
\emph{without}, lexical antonyms and pragmatic denial, none of which an explicit-cue list
catches. The demonstrated claim is therefore narrower than ``negation is a closed
class'': explicit negation cues in these annotation schemes are substantially more
enumerable than hedge and attribution realizations, and that is what predicts where
vocabulary repays effort. We report a rule that held wherever we could test it, not a
law; the way to break or extend it is to predict, in advance, how modality, quantifier
scope, evidentiality and reported-speech verbs behave, and then measure them.

\paragraph{The real-write results are a pilot.} \S\ref{sec:writes} and \S\ref{sec:mem0}
rest on $101$ blind labels, $29$ bad writes, $8$ scored production writes and one
extraction model. They are preliminary evidence, sized to establish direction and to
bound where the failure lives, not to estimate rates precisely. Every magnitude in them
should be read with the interval and the label-noise bound below attached.

\paragraph{One labeller, and the noise is now measured.} Every figure in
\S\ref{sec:writes} rests on $101$ judgements from a single annotator, who is also an
author. Self-agreement from a blind second pass is $70\%$ raw ($\kappa = 0.47$; on
flag/pass decisions alone, $77\%$, $\kappa = 0.55$), which is moderate, and it bounds
every number the corpus supports: magnitudes in \S\ref{sec:writes} should be read as one
careful but noisy reading, and only the direction claims (which mechanism dominates in
which domain) are stable under label noise of this size. Inter-annotator agreement is
not available.

\paragraph{The rubric is broader than the tool.} Labellers were asked whether a memory
would mislead a later reader, which is the right question about a memory and a wider one
than these seven checks implement. That is why \S\ref{sec:baserate} separates mechanisms
before reporting anything, and why no recall figure is given for the email corpus. An
earlier version of this analysis reported a single recall number against the broad
criterion; it was measuring the rubric.

\paragraph{Two corpora, several differences.} The domain comparison holds source
domain, labelling protocol and construction constant only in the first. The direction is
significant; the effect size is not clean, and we do not quote a ratio.

\paragraph{The pool is one extractor and one prompt.} Different memory systems consolidate
differently, and a system prompted to preserve stance launders less
\citep{kwon2026mc}. The base rates in Table~\ref{tab:baserate} are properties of a
pipeline, not constants of a domain.

\paragraph{Scope of the checks.} Detector-side bounds are in \S\ref{sec:limits}; in
summary: English only, substring-matched, paraphrase defeats it, broadening needs
ontology and is not attempted, and no external user has yet run the tool against a store
we did not construct.

\paragraph{The \textsc{added} numbers inherit a construct mismatch.} Its ground truth is
the corpus's ``out of scope'' mechanism, which was labelled against the broad
would-a-reader-be-misled rubric and therefore includes \emph{inferred} claims, while the
detector judges entailment against the source. Some of the recall gap is that seam
rather than detector error, and the single-labeller noise above bounds these figures
too. Its verdicts are also aggregated to the write from sentence-level answers.

\paragraph{The production score is one run of one system.} \S\ref{sec:mem0} is a single
pass over $15$ sources with one extraction model, and extraction is sampled: the same
stimulus produced no write in one run and this paper's sharpest laundering example in
another. Treat $5$ of $8$ as a measurement of that configuration on that stimulus set,
not as a property of the software, and note it is a flag rate: the gate's own false
alarms and its bounded recall move it in opposite directions.

\paragraph{Cost figures are one provider at one time.} The witness numbers use a small
model at 2026 prices and will not transfer.

\section*{Ethics Statement}
The Enron corpus \citep{klimt2004enron} is public correspondence from real people who did
not consent to its research use, and it is standard in NLP for that reason and in spite of
it. We mask email addresses and telephone numbers in every derived artifact. Personal
names are retained, because a relayed claim is unreadable without knowing who relayed it
and the failure under study is precisely the loss of that information. No corpus content
is redistributed: the released code downloads the archive and reproduces the pool locally.
Excerpts quoted in this paper were checked individually for personal content.

The tool is defensive. It examines text a system is about to store about its user and
reports what the compression dropped. It transmits nothing by default: the deterministic
path is entirely local, and the optional witness sends one sentence at a time to a
provider only when explicitly enabled.

\bibliography{references}

\appendix

\clearpage
\onecolumn

\noindent\begin{minipage}{\textwidth}
\subsection*{Appendix contents}
\noindent
\hyperref[app:claims]{\textbf{A}\quad Claims and evidence} \dotfill \pageref{app:claims}\\
\hyperref[app:sets]{\textbf{B}\quad Every evaluation set in one place} \dotfill \pageref{app:sets}\\
\hyperref[app:corpora]{\textbf{C}\quad Corpora and thresholds} \dotfill \pageref{app:corpora}\\
\hyperref[app:frank]{\textbf{D}\quad The whole gate on FRANK} \dotfill \pageref{app:frank}\\
\hyperref[app:prompts]{\textbf{E}\quad The witness prompt} \dotfill \pageref{app:prompts}\\
\hyperref[app:repro]{\textbf{F}\quad Reproducibility} \dotfill \pageref{app:repro}
\end{minipage}
\bigskip

\section{Claims and evidence}
\label{app:claims}

Every load-bearing claim, its evidence, and its epistemic status.

\begin{itemize}\itemsep2pt
\item \textsc{shown}: direct measurement supports it.
\item \textsc{retracted}: \emph{we} asserted it earlier and later withdrew it.
\item \textsc{not shown}: our measurement neither supports nor refutes it.
\item \textsc{not claimed}: we never asserted it; the row exists so a reader cannot
      infer it.
\end{itemize}

\medskip
{\small
\begin{longtable}{@{}L{0.46\textwidth} L{0.18\textwidth} L{0.28\textwidth}@{}}
\caption{\textbf{Claims and evidence.}}\label{tab:claims}\\
\toprule
\textbf{Claim} & \textbf{Evidence} & \textbf{Status} \\
\midrule
\endfirsthead
\multicolumn{3}{@{}l}{\emph{Table~\ref{tab:claims}, continued}}\\
\toprule
\textbf{Claim} & \textbf{Evidence} & \textbf{Status} \\
\midrule
\endhead
\midrule \multicolumn{3}{r@{}}{\emph{continued on next page}}\\ \endfoot
\bottomrule \endlastfoot

Closed-class negation detection transfers to untuned domains at $0.91$ F1. &
Tab.~\ref{tab:external} & \textsc{shown}, held-out documents, independently annotated
corpora \\

Open-class hedging and attribution plateau near half recall. &
Tab.~\ref{tab:external} & \textsc{shown} \\

The closed-class transfer prediction holds for \textsc{temporal}. & \S\ref{sec:external}
& \textsc{not shown}: no external corpus here annotates temporal scope; only its
false-positive rate is bounded, on the should-pass corpus \\

A witness recovers recall on open classes at equal precision. &
Tab.~\ref{tab:witness} & \textsc{shown}, $+17$ and $+15$ points \\

Vocabulary can close the open-class gap. &
\S\ref{sec:split} & \textsc{not shown}: three rounds of external mining moved held-out
adversarial recall $14\%\to29\%$, two cases of fourteen \\

\textbf{A witness allowed to raise verdicts closes the recall gap.} &
\S\ref{sec:limits} & \textsc{retracted}: our own claim. $93\%$ on the adversarial corpus,
zero recall gained on real writes and $7$ points of precision lost \\

Scope expansion is detectable without ontology. &
\S\ref{sec:limits} & \textsc{not shown}: the obvious signal fires on $86$ of $89$
should-pass items \\

The targeted failure is $55\%$ of bad writes in conversational hearsay and $7\%$ in
business email. & Tab.~\ref{tab:baserate} & \textsc{shown}: Fisher exact $p=0.00053$,
direction only; the corpora differ in more than domain \\

The $55\%$ figure estimates natural conversation. & \S\ref{sec:baserate} &
\textsc{not claimed}: those sources were constructed to contain hearsay, so it is an
upper bound for that setting \\

The gate reaches $0.34$ precision on business email. & \S\ref{sec:writes} &
\textsc{shown}, $n=47$ flagged writes labelled blind \\

\textbf{That precision figure measures the detector.} & \S\ref{sec:writes} &
\textsc{retracted}: our own first reading. With the base rate in
Tab.~\ref{tab:baserate} it is mostly a property of the domain \\

Recall on the failures the checks target, on business email. & \S\ref{sec:writes} &
\textsc{not shown}: the in-scope denominator is $2$; no ratio is reported \\

\textbf{A single recall figure against ``would a reader be misled'' measures this gate.} &
\S\ref{sec:limits} & \textsc{retracted}: our own analysis. That criterion includes
fabricated claims the checks cannot see, and reporting it measured the rubric \\

Narrowing \textsc{condition} to one sentence improves it, as it did for negation and
attribution. & \S\ref{sec:writes} & \textsc{not shown}: tried and measured worse at the
whole-gate level on the blind corpus (gate precision $0.34\to0.31$, estimated recall
$0.33\to0.21$; the baseline pair is \S\ref{sec:writes}'s own headline, since the variant
reruns the same scoring); the wide window catches real conditions \\

A direct LLM judge is the better detector on real memory writes. &
\S\ref{sec:judge} & \textsc{not shown}: it reaches $0.30$ (small) and $0.50$ (large)
recall against the gate's $0.95$ on the same $44$ writes, at higher precision; one
rubric, two models \\

\textbf{A judge cannot point at evidence.} & \S\ref{sec:judge} &
\textsc{not claimed}: it volunteered a source-grounded quote on $89$ of $90$ items. The
argument is that verdicts must be \emph{required} to cite text, not that models cannot \\

\textbf{Embedding alignment would raise real-world recall substantially.} &
\S\ref{sec:limits} & \textsc{not shown}: attribution of $12$ known misses gives at most
$1$ to the matcher; $10$ are lexicon losses on correctly aligned sentences \\

An \textsc{added} detector catches claims the source never supported. &
\S\ref{sec:tool} & \textsc{shown}: precision $0.57$ ($8/14$), recall $0.30$ ($8/27$),
coverage $96.9\%$ on the blind corpus, untuned; wide intervals at these denominators;
additive over a gate that catches none of this class \\

The gate flags $5$ of $8$ of \texttt{mem0} $2.0.7$'s hedged-hearsay writes. &
\S\ref{sec:mem0} & \textsc{shown}: one stimulus set, one extraction model, single run \\

\textbf{That is \texttt{mem0}'s laundering rate.} & \S\ref{sec:mem0} &
\textsc{not claimed}: a flag rate errs both ways (bounded recall hides cases; $\sim$1
flag in 4 is a false alarm), and $5/8$ carries a $95\%$ interval of roughly
$[0.31, 0.86]$ \\

\textbf{$8$ of $13$ real \texttt{mem0} writes fabricate.} & \S\ref{sec:mem0} &
\textsc{not claimed}: \textsc{added} fires on $8$, but $5$ are timestamp resolution and
$1$ an attribution shift; only $2$ are invention with no source anchor \\

An in-lexicon chain of rewrites can launder gradually past a per-write gate. &
\S\ref{sec:tool} & \textsc{not shown}: the gate composes; the hop dropping a class's
last cue fires. Chains escape by losing checkability instead \\

A value-preserving unit change is detectable where the stance checks pass. &
\S\ref{sec:tool} & \textsc{shown}: ``1.2 million dollars'' stored as ``1.2 million
euros'' yields \textsc{changed}/fail while the gate passes; pinned by shipped tests,
with the same-entity restriction and its known false negative documented \\

\textbf{Label noise is measured, and it is moderate, not small.} & \S\ref{sec:writes} &
\textsc{shown}: blind second pass, $70\%$ raw ($\kappa=0.47$), $77\%$
($\kappa=0.55$) on flag/pass alone; the second pass was stricter ($6{:}2$
pass$\to$flag). Every magnitude in \S\ref{sec:writes} inherits this bound \\

\end{longtable}
}

\twocolumn

\section{Every evaluation set in one place}
\label{app:sets}

This paper reports numbers from eight different sets, and two of them are precisions
that look contradictory until you know which is which: $0.73$ is the gate's precision on
the live-run calibration corpus, and $0.34$ is its precision on the blind Enron writes,
where \S\ref{sec:baserate} shows the targeted failure is rare. Table~\ref{tab:sets}
gives each set once, with what it measures and what it cannot.

\begin{table*}[t]
\centering\small
\begin{tabular}{@{}L{0.19\textwidth} L{0.11\textwidth} L{0.31\textwidth} L{0.31\textwidth}@{}}
\toprule
\textbf{Set} & \textbf{Size} & \textbf{What it measures} & \textbf{What it cannot} \\
\midrule
BioScope + Szeged + PolNeAR (test halves) & $105{,}596$ sents &
Cue detection per property; the transfer result of Tab.~\ref{tab:external} &
Nothing relational: no source/memory pairs, no temporal gold \\
\addlinespace
Witness-vs-lexicon gold & $200$/task &
Detector-level hedge and attribution recall, stratified $50/50$ &
Gate verdicts; the shipped witness only lowers them \\
\addlinespace
FRANK & $4{,}800$ sents &
Whole-gate agreement with typed human error labels (App.~\ref{app:frank}) &
Our failure: its typology is dominated by hallucination \\
\addlinespace
Live-run calibration & $109$ writes &
Gate precision $0.73$, recall $0.95$; the $0.25$ containment threshold &
Blind labelling; it is the corpus the thresholds were swept on \\
\addlinespace
Blind Enron writes & $101$ labelled ($94$ usable) &
Gate precision $0.34$, the $55/7$ mechanism split, label noise $\kappa=0.47$ &
Recall on in-scope failures: the denominator is $2$ \\
\addlinespace
\textsc{added} evaluation & $92$ writes &
Precision $0.57$, recall $0.30$, coverage $96.9\%$ &
Entailment vs.\ inference: the gold includes inferred claims \\
\addlinespace
Adversarial (visible + held-out) & $29 + 22$ items &
The memorisation gap, $92\%$ vs $14\%$, and the $29\%$ ceiling &
Natural prevalence; both sets were written to defeat a lexicon \\
\addlinespace
\texttt{hearsay-v1} production run & $15$ sources, $13$ writes &
\texttt{mem0} flag rate $5/8$ hearsay, $1/5$ legit (\S\ref{sec:mem0}) &
Anything general: one system, one run, sampled extraction \\
\bottomrule
\end{tabular}
\caption{\textbf{Every evaluation set in this paper, once.} The two precisions that look
contradictory are different corpora: $0.73$ is the live-run calibration set, $0.34$ the
blind Enron writes.}
\label{tab:sets}
\end{table*}

Three sets describe extractor output and are routinely confused: the $109$-write
live-run corpus (thresholds), the $89$-item hand-built should-pass corpus (false-positive
bounds for new checks), and the $101$ blind Enron writes (the only blind-labelled one).
They are disjoint in construction and in purpose.

\section{Corpora and thresholds}
\label{app:corpora}

\paragraph{Corpora.} BioScope and the Szeged Uncertainty Corpus contribute $57{,}891$
sentences of speculation and negation cues over biomedical abstracts, full papers,
encyclopedic text and news; PolNeAR contributes $47{,}705$ sentences ($1{,}008$
political news articles) with token-level source/cue/content attribution spans and its
own train/dev/test split, which we use as shipped. Szeged's \emph{investigation} subtype
(``we examined whether X'') is excluded and counted: it marks research framing, a real
uncertainty cue in a paper and an irrelevant one in a memory write.

One exclusion is worth recording because it looked like a result. Scoring BioScope's
public clinical file produced a clean $0.00$ precision across $6{,}383$ sentences, which
turned out to be a property of the release rather than the detector: every token in that
distribution of the clinical subcorpus is redacted to \texttt{*}. A harness bug that
deflates looks like rigour; it was caught only because the number was too clean.

\paragraph{Mining rule.} Fixed before looking: a lexicon candidate mined from dev is
kept only if the sentences it newly fires on are at least $60\%$ gold-annotated and it
fires at least ten times. Documents are split in half; every reported number is from
test-half documents disjoint from the tuned-on half.

\paragraph{Windows.} Which source context a check reads is measured, not assumed:
hedging is read from the matched sentence plus its neighbours because hedges float
across sentence boundaries (``Alice has admin. Not sure though.''); negation and
attribution are read from the matched sentence alone because they attach to their
clause. The narrowing was earned on FRANK, where wide windows read cues from
neighbouring sentences, and is pinned by a regression test. The claim-alignment
containment threshold is $0.25$, chosen by sweeping $109$ real extractor outputs (the
live-run calibration corpus, a third set distinct from both the $89$-item should-pass
corpus and the blind Enron writes): false positives are flat from $0.20$ to $0.50$
while recall falls as the threshold rises.

\section{The whole gate on FRANK}
\label{app:frank}

The two defects \S\ref{sec:frank} reports, both invisible to every corpus this project
built. First, the checks for clause-attached properties (negation, attribution) were
reading cues from neighbouring sentences, so a ``not'' next door denied a claim it had
nothing to do with. Second, the memory-side list held inflected forms with no stems, so
a memory reading ``german media \emph{say}'' was blocked while ``says'' would have
passed. Fixing both took the clean-sentence flag rate from $65\%$ to $37.5\%$ while
raising the lift on circumstance errors, the one FRANK type these checks target, from
$1.25\times$ to $1.59\times$ base rate: the gate fires less and discriminates better.
Real-world precision moved $68\%\to73\%$ with recall unchanged at $95\%$.

\section{The witness prompt}
\label{app:prompts}

The witness system prompt, verbatim:

\begin{promptbox}
You label ONE sentence for epistemic stance. You do not judge, compare, or decide
anything.\\[4pt]
Answer two questions about the sentence you are given:\\[4pt]
hedged: does the sentence mark the claim as uncertain, tentative, unofficial,
provisional, second-hand, or otherwise not established as fact?\\[4pt]
attributed: does the sentence indicate the claim came from some person, group, channel,
or document, rather than asserting it flatly on the speaker's own authority?\\[4pt]
A sentence can be both, either, or neither. Judge only what the sentence itself says, by
its meaning rather than by any fixed list of words. Ordinary domain vocabulary is not
stance: a confident statement stays confident even when it happens to contain words that
look epistemic in other contexts.\\[4pt]
Reply with ONLY this JSON and nothing else:\\
\{"hedged": true/false, "attributed": true/false, "markers": ["the exact words that made
you say yes"]\}
\end{promptbox}

The prompt deliberately contains no example idioms. An earlier version listed exact
phrasings from the held-out corpus, because it was written while looking at the
failures; it scored $100\%$, which measured the leak rather than the model. Stripping
the examples gave the honest $93\%$. Markers returned by the model are validated against
the sentence before use, and a reply that fails validation is treated as unusable: the
deterministic verdict stands.

\section{Reproducibility}
\label{app:repro}

\paragraph{Models.} Every model-assisted number here comes from one of two. Claude Haiku
4.5 \citep{anthropic2025haiku} runs the stance witness, the \textsc{added} detector, the
small judge of \S\ref{sec:judge}, and \texttt{mem0}'s own extraction in
\S\ref{sec:mem0}; Claude Sonnet 5 \citep{anthropic2025sonnet} runs the large judge. Both
were called at defaults, with extended thinking disabled where the API allows it, since
none of these are reasoning tasks and a production extractor would not pay for one. The
deterministic gate uses no model at all, which is why the suite runs offline.

The repository is \url{https://github.com/collapseindex/factwash} (Apache-2.0). The full
test suite runs offline with no API key, including the external-corpus evaluations; the
corpora download scripts fetch only freely available data. Metered API spend across the
experiments is \$$1.02$; the \textsc{added} evaluation and the production score of
\S\ref{sec:mem0} were run outside that harness and cost roughly \$$0.15$ more, which is
an estimate rather than a ledger figure. Metered runs are resumable and budget-capped,
with each guarantee broken on purpose by a test (including a simulated kill mid-write). Outputs follow a timestamped naming convention carrying operation,
model, parameters and seed, so lineage is recoverable from a filename alone.

Every figure published in the project README is recomputed from the shipped corpora by a
test that fails when the text drifts from the measurement, and that guard is
negative-tested: breaking a lexicon term or reverting a window makes it name the drift.
The claims ledger of Appendix~\ref{app:claims} is the same discipline applied to this
document.

\end{document}